\documentclass[a4paper,fleqn]{cas-dc}

\usepackage[numbers]{natbib}
\usepackage{algorithm,algorithmic,array,amssymb,amsthm,amsmath,amstext,booktabs,color,cite,float,graphicx,soul,multirow,threeparttable,subfigure,bm}
\usepackage{flushend}
\definecolor{hl}{rgb}{0.75,0.75,0.75}
\sethlcolor{hl}

\def\tsc#1{\csdef{#1}{\textsc{\lowercase{#1}}\xspace}}
\tsc{WGM}
\tsc{QE}
\tsc{EP}
\tsc{PMS}
\tsc{BEC}
\tsc{DE}

\begin{document}
\let\WriteBookmarks\relax
\def\floatpagepagefraction{1}
\def\textpagefraction{.001}
\shorttitle{}

\title [mode = title]{Attention-Guided Black-box Adversarial Attacks with Large-Scale Multiobjective Evolutionary Optimization}
%
%

\address[1]{Anhui Provincial Key Laboratory of Multimodal Cognitive Computation,  Anhui University, Hefei, Anhui 230601, CHN}
\address[2]{Institute of Multimedia and AI Security, School of Computer Science and Technology, Anhui University,  Hefei, Anhui 230601, CHN}
\address[3]{School of Computer and Information, Anqing Normal University, Anqing, Anhui 246133, CHN}

\author[1,2]{Jie Wang}

\author[1,2]{Zhaoxia Yin}[orcid=0000-0003-0387-4806]
\cormark[1]
\ead{yinzhaoxia@ahu.edu.cn}

\author[3]{Jing Jiang}

\author[1,2]{Yang Du}

\cortext[cor1]{Corresponding author}
%

\begin{abstract}
Fooling deep neural networks (DNNs) with the black-box optimization has become a popular adversarial attack fashion, as the structural prior knowledge of DNNs is always unknown. Nevertheless, recent black-box adversarial attacks may struggle to balance their attack ability and visual quality of the generated adversarial examples (AEs) in tackling high-resolution images. In this paper, we propose an attention-guided black-box adversarial attack based on the large-scale multiobjective evolutionary optimization, termed as LMOA. By considering the spatial semantic information of images, we firstly take advantage of the attention map to determine the perturbed pixels. Instead of attacking the entire image,  reducing the perturbed pixels with the attention mechanism can help to avoid the notorious curse of dimensionality and thereby improves the performance of attacking. Secondly, a large-scale multiobjective evolutionary algorithm  is employed to traverse the reduced pixels in the salient region. Benefiting from its characteristics, the generated AEs have the potential to fool target DNNs while being imperceptible by the human vision.  Extensive experimental results have verified the effectiveness of the proposed LMOA on the ImageNet dataset. More importantly, it is more competitive to generate high-resolution AEs with better visual quality compared with the existing black-box adversarial attacks.
\end{abstract}

%


\maketitle

\section{Introduction}
Recent years have witnessed the emergence and explosion of deep learning, since deep neural networks (DNNs) have been widely applied in various fields, such as computer vision, natural language processing, and speech recognition \citep{vaswani2017attention,xie2019exploring,abate2020deep}. Especially in the field of image recognition, the top-5 predictive error against ImageNet is decreased steadily from 16.4\% to 2.25\% during the last several years \citep{Hu2019SeNet,krizhevsky2012imagenet}. The evidence shows that the robustness of DNNs has been improved rapidly through effective neural architecture design and regularization techniques. Therefore, there is now a consensus that using DNNs is more suitable than the other models for image recognition and its relevant applications.

In the past decade, a series of studies have shown that DNNs are vulnerable to adversarial examples (AEs) by imposing some designed perturbations to original images \citep{szegedy2013intriguing,goodfellow2014explaining,moosavi2016deepfool,carlini2017towards,chen2021finefool,guoelaa2021}. These perturbations are imperceptible to human beings but can easily fool DNNs, which raises invisible threats to the vision-based automatic decision \citep{kurakin2016adversarial,Yin2020,ye2020detection}. Consequently, the robustness of DNNs encounters great challenges in real-world applications. For example, the existence of AEs can pose severe security threats for the traffic sign recognition in the autonomous driving \citep{bian2021adversarial}. AEs in the object detection will also influence the region proposals and bring undesirable segmentation and classification \citep{wang2020adversarial}.  Based on above evidence, the issue of AEs has received considerable attention \citep{zhang2019adversarial,martin2020inspecting}.

Szegedy \emph{et al.} first pointed out the vulnerability of DNNs and proposed the definition of adversarial attacks \citep{szegedy2013intriguing}.  They also demonstrated that the AEs for one network could fool another, even DNNs were trained on different datasets. Then, a considerable amount of researches on adversarial attacks has been studied. These attacks are designed to fool the target DNN by adding a small perturbation $\bm{X}$ to the original image $\bm{I}$ : $\mathcal{C}(\bm{I} + \bm{X}) \ne \mathcal{C}(\bm{I})$, where $\mathcal{C}( \cdot )$ is a $ m $-class classifier that receives $ n $-dimensional input and gives $ m $-dimensional output. AEs can be easily generated by using internal information of the target DNN, \emph{e.g.,} the gradient of the loss function of the original image \citep{goodfellow2014generative,kurakin2016adversarial1,rozsa2016adversarial,madry2017towards}. These attacks, called white-box attacks, are essentially an exploration of the robustness of DNNs. Other than the white-box attacks, researchers have shown an increased interest in black-box attacks. To be specific, the attacker can only obtain the output of the target DNN without accessing its structures and parameters. Since the structural prior knowledge of DNNs is usually unavailable, the works on black-box attacks are more practical than that of white-box ones in real cases. Therefore, numerous attempts have been made to realize black-box attacks, such as the gradient estimation-based \citep{tu2019autozoom}, query-based \citep{brendel2017decision,chen2017zoo}, or transferability of AEs-based \citep{dong2019evading} attacks.

To date, several works suggest that there is more than one objective should be taken into consideration in attacking, \emph{e.g.,} minimizing the confidence probability of the true label and the perturbation intensity of the changed image simultaneously \citep{liu2019black,suzuki2019adversarial}. They expected that the candidate AEs would mislead the target DNN as possible while exhibiting similar visual features with the original image. Unfortunately, these two objectives are somewhat conflicting with each other. On the one hand, the great perturbations significantly influence the classification result of DNNs, but the generated AEs could be easily detected by the human vision, as they lose the majority of features of the original image. On the other hand, a slight change that is hardly observed by both human beings and computers is not enough to mislead DNNs. Most of the previous methods cannot balance these two objectives and thereby probably miss the optimal trade-off perturbation. The issue stimulates the efforts on evolutionary algorithm-based, especially multiobjective evolutionary algorithm-based black-box adversarial attacks. Moreover, since the traditional evolutionary algorithm are not competive on handling the optimization problems with large-scale decision variables \citep{Yang2008L}, the existing works following this line may not achieve satisfactory performance on the high-resolution images.

\begin{figure}
	\centering
	\includegraphics[width=3.5in]{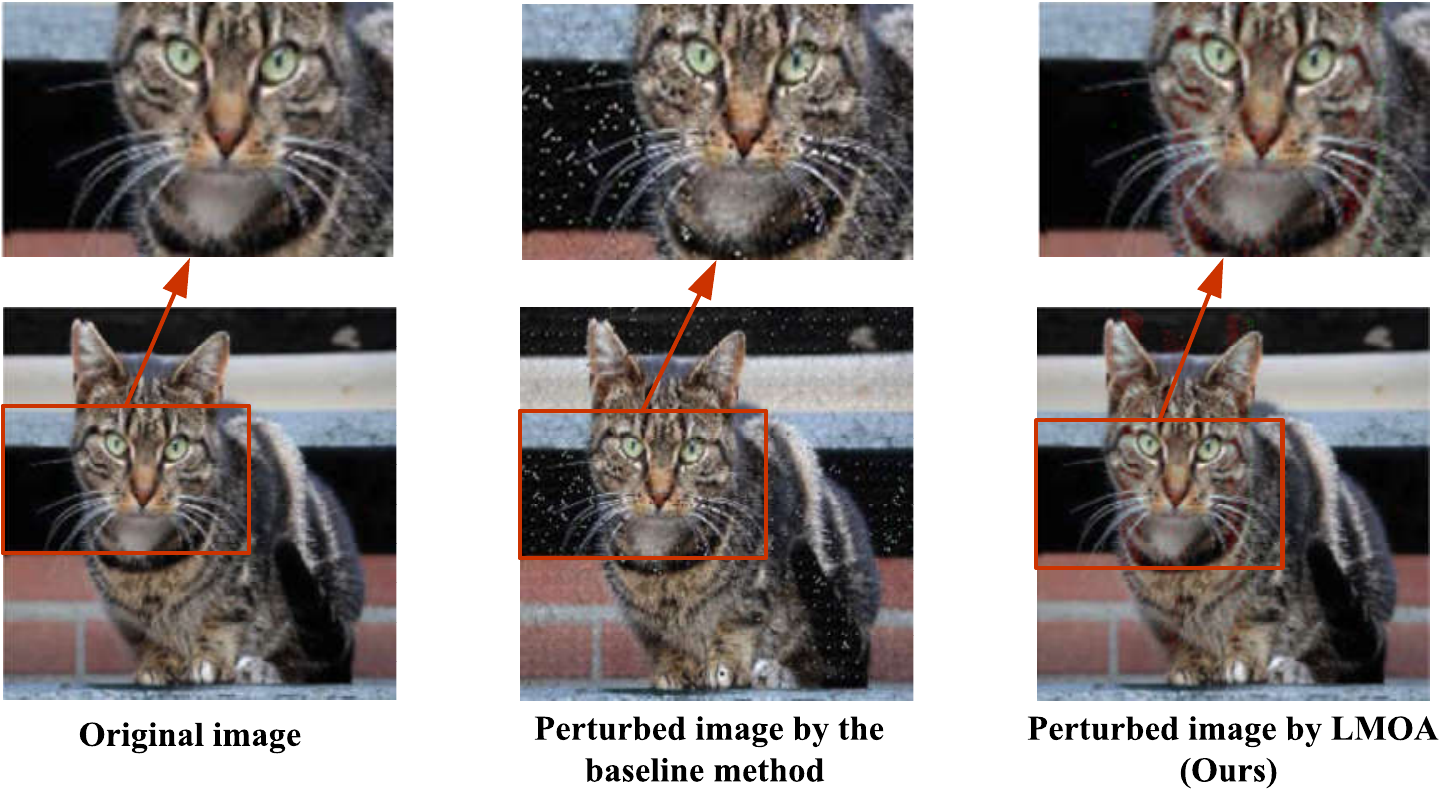}
	\caption{The original image and its corresponding adversarial examples generated by the MOEA-based attack \citep{suzuki2019adversarial} and LMOA, respectively.}
	\label{fig:1}
\end{figure}

To overcome the above drawbacks, we propose an atten\\tion-guided black-box adversarial attack, where a large-scale multiobjective evolutionary algorithm is employed to traverse the salient region of an image. Since the proposed method involves \underline{L}arge-scale \underline{M}ultiobjective \underline{O}ptimization and \underline{A}ttentional mechanism, it is named by LMOA. The main contributions of this study can be summarized as follows:
\begin{enumerate}
	\item \textbf{Using the attention mechanism to screen the attack-ed pixels.} We firstly use the class activation mapping (CAM) and a proxy model to obtain the attention map of the target image. The map strictly limits the attacked pixels so that the perturbations are only allowed to emerge within the salient region. On the one hand, attacking the salient pixels might be more efficient than the entire image, as these pixels can better reflect the spatial semantic information. On the other hand, screening the pixels is able to reduce the dimensionality of decision variables in the case of high-resolution images, which is beneficial to the black-box optimization.
	\item \textbf{Performing the black-box adversarial attack with a large-scale multiobjective evolutionary algorithm.} We secondly formulate the black-box attack into a large-scale multiobjective optimization problem, in which both attack ability and visual quality of the generated AEs are viewed as two objectives. Then, an optimizer tailored for large-scale optimization is employed. By doing so, a set of Pareto optimal solutions that achieves the balance between two objectives will be obtained, and the final generated perturbations can easily fool the target DNNs while being imperceptible by the human vision.
	\item \textbf{Attacking high-resolution images with high success rate and acceptable visual quality.} Extensive experimental results have been investigated on the ImageNet dataset. The results show that the proposed LMOA can achieve almost 100\% success rate of attacks. Compared with the MOEA-based attack, LMOA is more competitive to contribute high-resolution AEs with better  visual imperceptibility (see Fig. \ref{fig:1}).
\end{enumerate}

The structure of the remaining contents is organized as follows. Section II introduces the related works and motivation of this paper. The methodology of the proposed method is given in Section III. The experimental studies and ablation study are presented in Section IV. Finally, this paper is concluded in Section V.

\section{Related work}
In this section, we summarize recent approaches for generating AEs, in both the white-box and black-box cases. Besides, we emphasize existing evolutionary algorithm based attacks. Please refer to the cited works for further details.

\subsection{White-box adversarial attacks}
In white-box attacks, the attacker is allowed to obtain the structure and parameters of the target DNNs. Among the studies, the gradient-based approach is the most commonly used white-box attack. Goodfellow \emph{et al}.\citep{goodfellow2014explaining} proposed a simple and fast attack, named fast gradient sign method (FGSM), where AEs are generated based on the classification loss gradient of the input image. In their following works, an iterative version of FGSM (IFGSM) is proposed to obtain better attack performance \citep{kurakin2016adversarial1}, whereas a momentum term was suggested in \citep{dong2018boosting} to improve the transferability of adversarial samples. Chen et al. \citep{chen2021finefool} proposed FineFool, a channel-space and pixel-space attention-based adversarial attack. By seeking the relationship between perturbation and object contours, FineFool provides a better attack performance with less perturbation. Guo et al. \citep{guoelaa2021}proposed a local attack method (ELAA) that uses a model interpretation approach to find the identification region, which is combined with the original adversarial attack to achieve.
Other than the above works, a great deal of work has also been presented, aiming at obtaining AEs that are more robust to unknown and more defensive DNNs \citep{dong2019evading,dong2020robust}. However, the structural prior information of DNNs is usually unknown in realistic cases. Therefore, many researchers focus on the research toward black-box attacks.

\subsection{Black-box adversarial attacks}
Black-box attacks assume that the attacker only knows the outputs of the target DNN, \emph{e.g.,} category labels or confidence scores. They are categorized into three main classes \citep{bhambri2019survey}: gradient estimation-based \citep{chen2017zoo,nitin2018practical,du2018towards}, transferability-based  \citep{papernot2017practical,shi2019curls,dong2019evading}, and query-based methods \citep{narodytska2017simple,alzantot2019genattack,guo2019simple}. The gradient estimation-based methods perform adversarial attacks by estimating the gradient of the target model. Although they can effectively generate AEs, these methods highly rely on the accuracy of the estimated gradient. As for the transferability-based methods, the attacker usually utilizes the transferability of the target DNN, and trains an alternative/surrogate model for adversarial attacks.  It is clear that this type of attack relaxes the necessity of the inner information of DNNs, but heavily depends on the transferability, which is sensitive to the mismatch between the alternative and target DNNs. The third class is the query-based methods, which follow a query-feedback pattern. In detail, a candidate AE begins as a random input and changes by the disturbance noise with an acceptable level.
Query-based attacks allow the attackers query the output of the target DNN based on the inputs. To be specific, Brendel et al. \citep{brendel2017decision} introduced a query-based attack, called Boundary Attack. Boundary Attack takes an irrelevant image or target image as a starting point and gradually reduces the interference, making the generated AE seems similar to the input image. Chen \emph{et al.} \citep{chen2017zoo} suggested a ZOO attack, where the loss function is modified to make itself only rely on the output of DNNs. The advantage of query-based methods is that they can accomplish adversarial attacks without considering the gradient information and alternative model. However, their performance is also limited by the dimension and direction of the search. The subsequent studies have revealed that they may not alleviate the computational cost as they often require a large number of queries \citep{alzantot2019genattack}. 

\subsection{Evolutionary algorithm based attacks}
Evolutionary algorithm (EA) is a computational model that simulates the natural selection and genetic mechanisms of Darwinian evolutionary biology \citep{fonseca1995overview,van2000multiobjective,Mukhopadhyay2014,Han2018}. The main features of EA are characterized as derivative-free and global search, enabling EA becomes a more suitable method for solving black-box adversarial attacks. Recently, Su \emph{et al}.\citep{su2019one} proposed an one-pixel attack based on differential evolution (DE) \citep{storn1997differential}, where only a limited number of pixels in the image are changed to construct AEs. Due to the usage of DE, this method does not require the gradient of the target DNN and can be integrated with a non-differentiated objective function. Note that, a few pixel changes cannot result in a significant decline of the classification confidence with large-scale pixels, such that the success rate of adversarial attacks is unfavorable in the case of high-resolution images. Liu \emph{et al}. \citep{liu2019black} presented a EA-based method that focuses on attacking the entire image, rather than limiting the number of changed pixels. The optimization objective is formulated as a weight function, where the minimization of confidence of the original label is considered as the main objective, while the similarity between solutions and the original image is used as a penalty.

The above EA-based approaches usually formulate a single-objective constrained/unconstrained optimization model. In essence, the process of AEs generation requires considering more than one objective, such as the classification probability and the visual quality of the candidate AEs. Therefore, the task of black-box adversarial attacks can be reformulated as a multiobjective optimization problem (MOPs). Mathematically, a MOP is stated as \citep{Deb2002A}
\begin{equation}\label{MOPd}
	\begin{array}{cl}
		\min  & \textbf{F}(\textbf{x}) = (f_{1}(\textbf{x}), f_{2}(\textbf{x}), \cdots, f_{m}(\textbf{x}))^{\texttt{T}} \\
		\mbox{s.t.} & \textbf{x} \in \Omega
	\end{array},
\end{equation}
where $\Omega$ is the \emph{decision space} and $\textbf{F}$ consists of $m$ objective functions, respectively. $\{\textbf{F}(\textbf{x})|\textbf{x} \in \Omega\}$ is the \emph{attainable objective set}. Considering two decision vectors $\textbf{x}_{1},\textbf{x}_{2} \in \Omega$, the vector $\textbf{x}_{1}$ is said to \emph{dominate} $\textbf{x}_{2}$ (denoted by $\textbf{x}_{1}\prec \textbf{x}_{2}$) if and only if $\forall i:f_{i}(\textbf{x}_{1})\leq f_{i}(\textbf{x}_{2})\, \wedge \exists j: f_{j}(\textbf{x}_{1})< f_{j}(\textbf{x}_{2})$, where $i, j \in \{1,\cdots,m\}$. Moreover, a solution $\textbf{x}^{\ast}$ is \emph{Pareto optimal}, if there is no other solution $\textbf{x} \in \Omega $ such that $\textbf{x} \prec \textbf{x}^{\ast}$. The set of all the Pareto optimal solutions is defined as Pareto set (PS) and its image is called Pareto front (PF).

By contrast with single-objective EAs, multiobjective evolutionary algorithms (MOEAs) are characterized as population-based metaheuristics for handling MOP \citep{Sun2020evolving,Jiang2020Eff,Xiang2020}, each of which has more than one Pareto optimal solution \citep{Zhou2011,Trivedi2017}. In this community, research efforts are devoted to discovering the tradeoff solutions, which are as close as possible to the Pareto optimal solutions and cover the Pareto front as uniformly as possible \citep{Cheng2016A,Deb2014An}. As for MOEA-based attacks, Suzuki \emph{et al}. \citep{suzuki2019adversarial} first demonstrated the feasibility of using MOEAs for black-box attacks in three scenarios. While this method can effective attack against low-dimensional images (\emph{e.g.}, CIFAR-10), its performance will degenerate when the dimension of the optimization problem rises. Consequently, it is necessary to apply dimension reduction techniques or MOEAs tailored for large-scale tasks to attack high-resolution images (e.g., ImageNet).

\subsection{Motivation}
Based on the above descriptions, there are still some limitations in conventional studies on EA-based black-box attacks. Therefore, the motivations of this paper are summarized in the following two respects.

\textbf{1) Screening the perturbed pixels can reduce the complexity of optimization while generating more effective AEs.} Previous adversarial attacks usually focus on attacking the entire image, regardless of its spatial semantic information. It is also worth noting that the foreground region usually contains more texture information than the background one of an image. In other words, changing the pixels in the foreground region of an image is more effective than attacking the entire image \citep{dong2020robust}. The fact motivated us to use an attention mechanism to restrict the attack within the salient pixels in the foreground region. More importantly, from the perspective of optimization, this implementation can greatly reduce the search space and the complexity of the optimization.

\textbf{2) Previous EA-based black-box attacks encounter difficulty in tackling high-resolution images.} Recent evidence suggests that the AEs generation can be formulated as a MOP. However, current EA-based approaches usually model the black-box attack as a constrained single-objective optimization problem or an unconstrained problem with weights or penalties. The studies cannot balance the solutions between two optimization objectives, and their optimization result is sensitive to the weights. Although several researchers have investigated the black-box attacks based on MOEAs, the performance of attacks will significantly degenerate as the dimension of the target image increases (see Fig. \ref{fig:1}). As for a high-resolution image, the dimension of decision variables to be optimized can reach hundreds of thousands or more, such that the complexity of the optimization greatly increases. In this case, using traditional MOEAs cannot obtain an acceptable convergence performance. The issue inspires us to use one of the MOEAs tailored for large-scale MOPs.

\begin{figure*}
	\centering
	\includegraphics[width=7in]{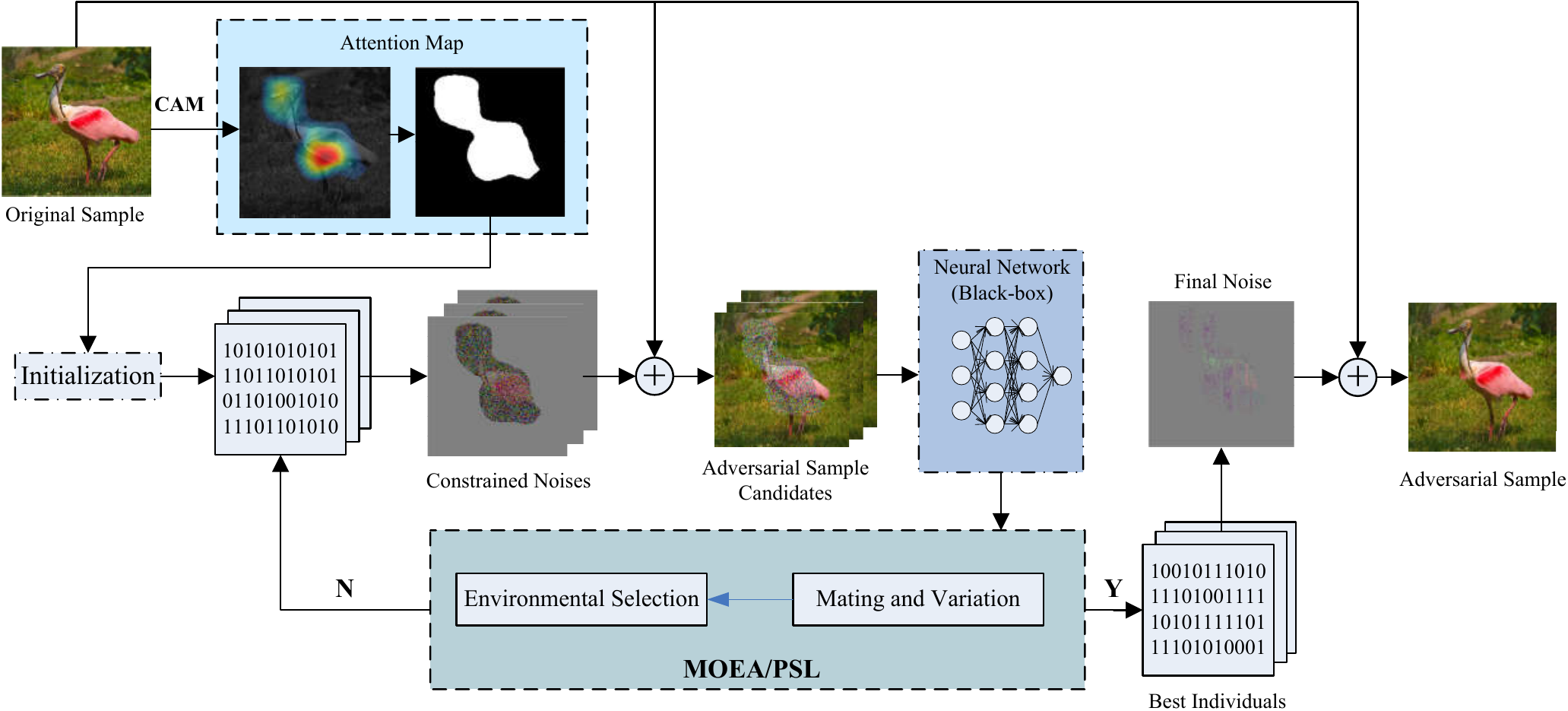}
	\caption{The general framework of the proposed LMOA.}
	\label{fig:2}
\end{figure*}

\section{Methodology of LMOA}
The general framework of the proposed LMOA is exhibited in Fig. \ref{fig:2}. LMOA mainly consists of two steps: 1)  Screening the perturbed pixels with the attention mechanism; 2) Generating the optimal perturbations with a large-scale MOEA. In the following subsections, we will detail these two steps.

\subsection{Screening perturbed pixels with the attention mechanism}
\begin{figure}[htbp]
	\centering
	\includegraphics[width=3.5in]{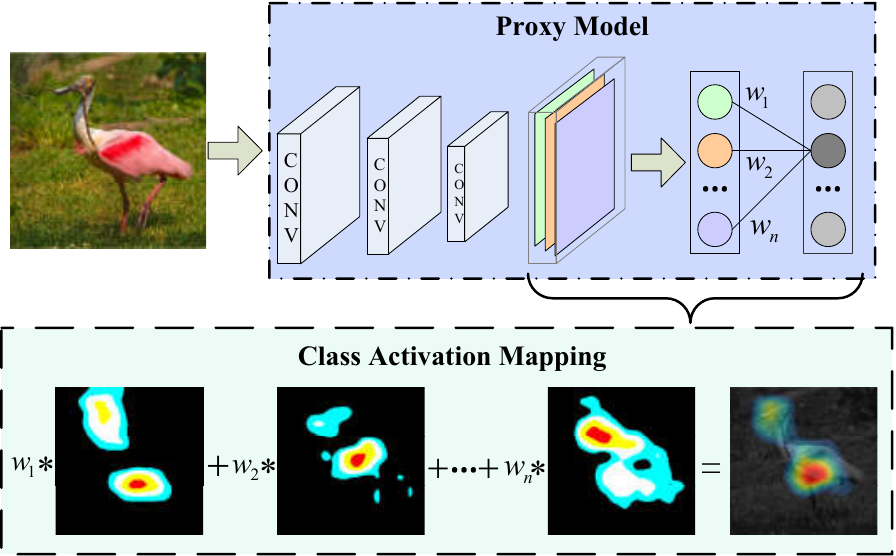}
	\caption{The process of obtaining CAM with a proxy model.}
	\label{fig:3}
\end{figure}

By considering the spatial semantic information, LMOA firstly employs the class activation mapping (CAM) \citep{zhou2016learning}  to obtain the attention map of the target image. CAM can visualize predicted class scores on any given image, highlighting the region of the object detected by the target DNN. In other words, the obtained attention map reflects the pixels of the interest of DNNs in the classification (as shown in Fig. \ref{fig:3}). Obviously, attacking these pixels that contain the spatial semantic information can fool the target DNN with a relatively higher probability.  However, it is tricky to know the gradient information of black-box DNNs, which brings a barrier for using CAM. Therefore, we suggest using a proxy model to obtain an approximated attention map of the input image.

To make the generated adversarial perturbations more effective, only the salient pixels derived from the attention map will be screened as the attacked pixels. To be specific, the attention map of the proxy model is binarized to represent the candidate pixels for attacking. The binarization can be formulate as
\begin{equation}\label{bina}
	{p_{i}} = \left\{
	\begin{array}{*{20}{c}}
		{0,{\rm{\quad}}{u_{i}} = 0}\\
		{1,{\rm{\quad}}{u_{i}} \ne 0}
	\end{array}\right.,
\end{equation}
Where $u_{i}$ is the value of pixel at the $(l,w,c)$ position of attention map (in detail, $l$ and $w$ determine the coordinate position of the pixel, and $c$ denotes the channel the pixel belongs to). Based on Eq. (\ref{bina}), the pixels with $p_{i} = 1$ is used to define the salient region of an image to be attacked. By doing this, the dimension of the decision variables to be optimized is reduced, as the pixels for perturbation is changed. The reduction of the search space also facilitates the convergence of the subsequent MOEA.

\subsection{Generating perturbations with large-scale MOEA}
The black-box attack is firstly formulated as a multiobjective optimization problem below:
\begin{equation}\label{MOP}
	\begin{array}{cl}
		\min  & {f_1} = P(\mathcal{C}(\bm{I} + \bm{X} ) = \mathcal{C}(\bm{I}))\\
		\min  & {f_2} = {\left\| \bm{X}  \right\|_0}\\
		\min  & {f_3} = {\left\| \bm{X}  \right\|_2}\\
		\mbox{s.t.} & 0 \le {u_{i}} + {x _{i}} \le 255
	\end{array},
\end{equation}
where $P(\cdot)$ denotes the \emph{confidence probability} of the classification result; $\bm{I}$ and $\bm{X}$ represent the \emph{original sample} and  \emph{adversarial perturbation}, respectively; $u_{i}$ is the value of pixel at the $(l,w,c)$ position of $\bm{I}$, while $x_{i}$ is the value of perturbation.

As shown in Eq. (\ref{MOP}), the proposed multiobjective optimization based black-box attack involves three objective functions. The first one ${f_1}$ represents the probability that the target classifier $\mathcal{C}( \cdot )$ classifies the generated adversarial example $\bm{I} + \bm{X}$ into the correct class $\mathcal{C}(\bm{I})$. The remaining two functions are both distance metrics, each of which is employed to evaluate the similarity between $\bm{I} + \bm{X}$  and $\bm{I}$. Furthermore, minimizing  ${l_0}$ distance (${l_0}$ norm) is to restrict the number of pixels to be attacked, while minimizing ${l_2}$ distance aims to reduce the change of each pixel. Other than three objective functions, the constraint imposed to $\bm{X}$ defines the range of perturbation on each pixel based on the intrinsic property of images. Using such a constraint can effectively reduce the search space and facilitate the convergence of MOEA.  Finally, we note that most of the decision variables in $\bm{X}$ are fixed as zero, and the dimension of perturbations to be optimized is relatively lower than that of $\bm{I}$. The reason is that the candidate attacked pixels are significantly reduced according to the technique introduced in the last subsection.

\begin{algorithm}[htbp]
	\caption{\emph{MOEA/PSL}}
	\begin{algorithmic}[1]\label{psl}
		\REQUIRE $N$ (population size)
		\ENSURE $P$ (final population)
		\STATE $P \leftarrow$ \emph{Initilization}($N$);
		\STATE Initialize the parameters $\rho$ and $K$;
		\WHILE{termination critertion not fullfilled}
		\STATE $P^{\prime} \leftarrow$ \emph{MatingSelection}($P$);
		\STATE $O \leftarrow$ \emph{Variation}($P$, $P^{\prime}$, $\rho$, $K$);
		\STATE $P \leftarrow$ \emph{EnvironmentalSelection}($P\cup O$);
		\STATE $[\rho, K] \leftarrow$ \emph{ParameterAdaption}($P$, $\rho$) ;
		\ENDWHILE
		\RETURN $P$
	\end{algorithmic}
\end{algorithm}

Note that, the objective $f_2$ is compatible with $f_3$ in some cases, \emph{e.g.}, a solution with a small objective value on $f_2$ may also has an acceptable performance on $f_3$. Moreover, the value of $f_2$ reflects the sparsity of a solution, that is, minimizing $f_2$ is to find the most sparse adversarial attack. We thereby reformulate the black-box attack below and resort to one of MOEAs tailored for large-scale sparse multiobjective optimization problems (LSMOPs).
\begin{equation}\label{SMOP}
	\begin{array}{cl}
		\min  & {f_1} = P(\mathcal{C}(\bm{I} + \bm{X} ) = \mathcal{C}(\bm{I}))\\
		\min  & {f_2} = {\left\| \bm{X}  \right\|_2}\\
		\mbox{s.t.} & 0 \le {u_{i}} + {x _{i}} \le 255
	\end{array}
\end{equation}

LSMOPs are characterized as the problems, where the value of most decision variables of their Pareto optimal solutions is zero, and the remaining large-scale variables are one or the other real numbers. Since recent methods can tackle this type of problems
\citep{Tian2020An}, we employ the MOEA based on Pareto-optimal subspace learning (MOEA/PSL), which is recently proposed in \citep{Tian2020So}, to solve the problem depicted in Eq. (\ref{SMOP}). In general, Algorithm \ref{psl} presents the framework of MOEA/PSL, which is very similar to the nondominated sorting genetic algorithm (NSGA-II) \citep{Deb2002A}. In the beginning, the population $P$ is initialized. It is worth noting that the candidate adversarial example is required to have a high similarity with the original image. Bearing this in mind, the population is randomly initialized at first, and then it is multiplied with a small number $\alpha$, \emph{e.g.,} $\alpha = 0.5$. Besides,  two parameters $\rho$ and $K$ that will be detailed in the following contents are also initialized. In the main loop, $N$ parents are selected according to their nondominated front numbers and crowding distances, through the binary tournament selection. Then, the offspring set $O$ is generated via the population $P$ and parent set $P^{\prime}$. Then, $N$ solutions with better nondominated front numbers and crowding distances survive from the combination of $O$ and $P^{\prime}$. Finally, the parameters $\rho$ and $K$ are adapted based on the new population $P$.

As for the representation, the length of each solution $\bm{x}$ is denoted by $D$, and each bit corresponds to one pixel.  For instance, given a high-solution image with a size of $(224,224,3)$, $D$ thereby equals to $224\times224\times3 = 150,528$ without using attention mechanism. To be specific, each solution $\bm{x}$ in $P$ is represented by a binary vector $\bm{xb}$ and a real vector $\bm{xr}$, and the $i\,$th dimension ($1 \le i \le D $) of $\bm{x}$ is obtained by
\begin{equation}
	x_i = xb_i \times xr_i ,
\end{equation}
where $xb_i$ indicates whether the perturbation on the pixel $u_{i}$ is zero, and $xr_i$ indicates the amount of perturbation on $u_{i}$. During the search, $\bm{xr}$ records the best perturbation on each pixel, while $\bm{xb}$ records the best positions that should be perturbed and controls the sparsity of adversarial perturbations.  Before the generation of offspring solutions, $\bm{xb}$s and $\bm{xr}$s of all the nondominated solutions are collected and used to train a denoising autoencoder (DAE) and a restricted Boltzmann machine (RBM), respectively. The number of hidden layers is controlled by the parameter $K$ ($K \ll D$) and iteratively updated by \emph{ParameterAdaption}, which is detailed in \citep{Tian2020So}. By doing so, the dimensions of $\bm{xb}$ and $\bm{xr}$ can be reduced via two models, and two reduced vectors are therefore generated. Moreover, the reduced vectors can also be recovered to the original vectors. In \emph{Variation}, two offspring solutions will be generated by simulated binary crossover (SBX) \citep{Deb2001Mu} and polynomial mutation (PM) \citep{Deb1996AC}, when the parameter $\rho$ is smaller than a random number. Otherwise, two parents will be fed into RBM and DAE to generate a pair of reduced vectors. Then, two new reduced vectors are accordingly obtained through SBX and PM, and they finally recovered as two offspring solutions. The update of the parameter $\rho$ can also refer to \citep{Tian2020So}. To summarize, the aim of using DAE and RBM is to learn the Pareto-optimal subspace, which enables MOEA/PSL to evolve in a low-dimensional search space.

\begin{figure*}
	\centering
	\includegraphics[width=\linewidth]{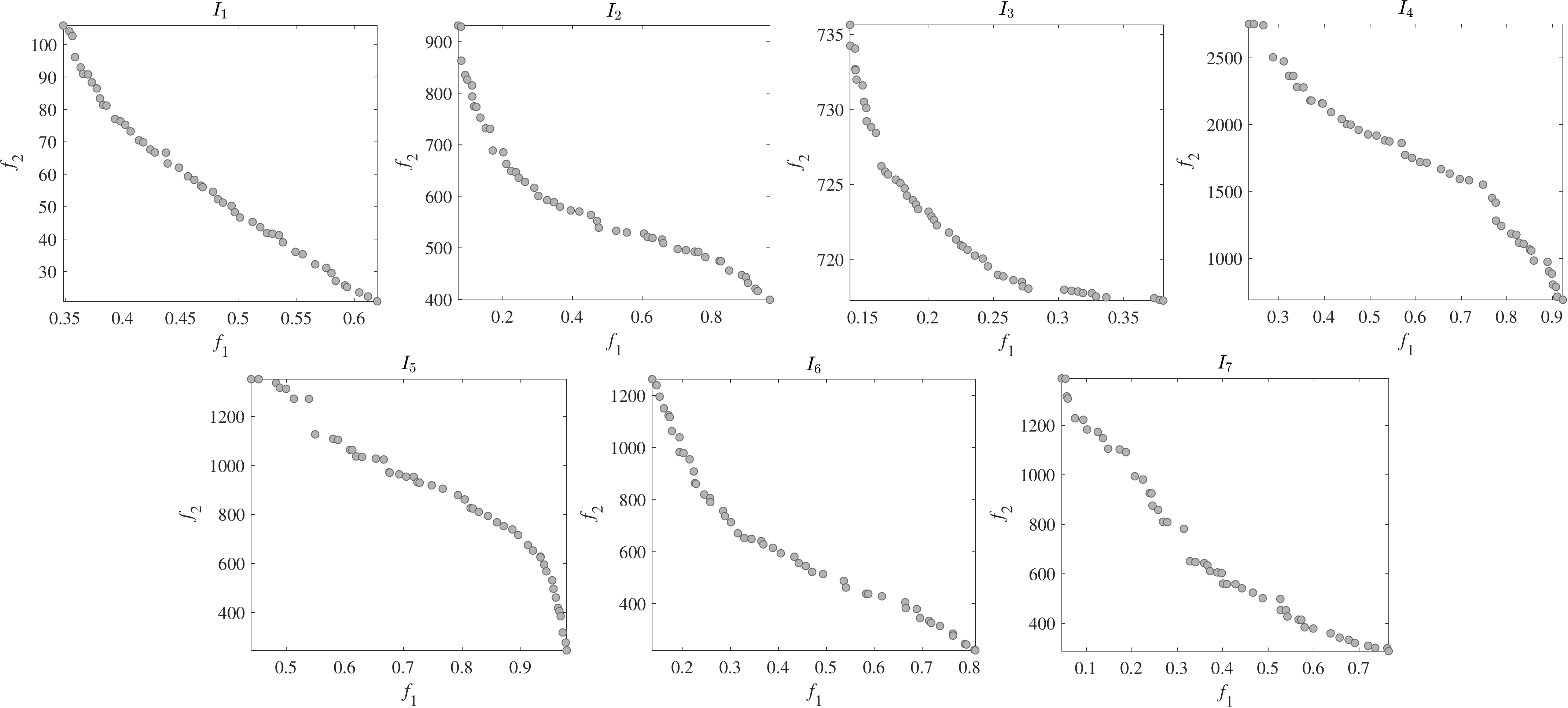}
	\caption{The Pareto optimal solutions obtained by LMOA with seven randomly selected images.}\label{Pareto}
	\label{fig:pareto}
\end{figure*} 

\section{Experiments and analysis}
In this section, we first introduce the experimental setup, including the benchmark dataset, parameter setting and baseline methods. Then, we investigate the performance of the proposed LMOA and compare it with state-of-the-art attacks, and MOEA-based attack\citep{suzuki2019adversarial}. Finally, the ablation study is performed to verify the effectiveness of using attention guided search and Large-scale MOEA.

\subsection{Experimental Setup}
\subsubsection{The benchmark dataset}
The benchmark dataset contains $2000$ high-resolution images that are randomly selected from ImageNet-1000 \citep{deng2009imagenet}, which consists of $1000$ categories in total. 

\subsubsection{The target and proxy models}
Two DNNs, including the pretrained ResNet-101 \citep{he2016deep} and Inception-v3 \citep{szegedy2016rethinking}, are selected as the target models for each compared algorithm. In addition, the SqueezeNet \citep{iandola2016squeezenet} is employed as the proxy model\footnote{Multiple proxy models have been tested and the results show the performance of LMOA is not sensitive to the type of proxy models} for obtaining the CAM in the proposed algorithm.

\subsubsection{Baseline methods} We compared LMOA with five different baseline methods, including three white-box attacks and two black-box attacks. Among them,  Fast Gradient Sign Model (FGSM) \citep{goodfellow2014explaining}, FineFool \citep{chen2021finefool} and ELAA \citep{guoelaa2021} are defined as the white-box attacks.  The remaining two methods, including Boundary Attack \citep{brendel2017decision} and ZOO \citep{chen2017zoo}  are classified into the black-box attacks.

\subsubsection{Parameters settings}
Before the optimization, the resolution of each image is resized according to the input layer of each model, \emph{i.e.,} $224\times224\times3$ for ResNet-101 and $299\times299\times3$ for Inception-v3, respectively.

\begin{table*}[htbp]
	\centering
	\renewcommand{\arraystretch}{1.5}
	\caption{Classification results and corresponding confidences of the original and AEs.}
	\begin{tabular}{c|cc|cc|cc|cc}
		\hline
		\multirow{3}{*}{Image No.}&   \multicolumn{8}{c}{Recognition results and confidence}\\ \cline{2-9}
		&   \multicolumn{4}{c|}{ResNet-101}&   \multicolumn{4}{c}{Inception-v3}\\ \cline{2-9}
		& \multicolumn{2}{c|}{$\mathcal{C}(\bm{I})$}&    \multicolumn{2}{c|}{$\mathcal{C}(\bm{I} + \bm{X})$}&   \multicolumn{2}{c|}{$\mathcal{C}(\bm{I})$}&  \multicolumn{2}{c}{$\mathcal{C}(\bm{I} + \bm{X})$}\\\hline
		$I_1$&         Vulture:&               62.05\%&         Kite:&             52.37\%&     Kite:&            15.30\%&     Kite:&            70.98\% \\
		$I_2$&         Hotdog:&                95.69\%&         Cucumber:&         47.46\%&     Banana:&          10.03\%&     Banana:&          59.93\% \\
		$I_3$&         Street sign:&           50.28\%&         Shopping cart:&    57.47\%&     Street sign:&     15.19\%&     Shopping cart:&   88.76\% \\
		$I_4$&         Wolf spider:&           92.69\%&         Tarantula:&        51.47\%&     Wolf spider:&     93.70\%&     Barn spider:&     43.05\%\\
		$I_5$&         Knot:&                  97.67\%&         Swab:&             53.19\%&     Knot:&            92.69\%&     Swab:&            42.84\%\\
		$I_6$&         Fig:&                   81.77\%&         Jackfruit:&        22.90\%&     Fig:&             99.06\%&     Mushroom:&        44.82\%\\
		$I_7$&         Hip:&                   77.13\%&         pomegranate:&      42.94\%&     Hip:&             33.66\%&     Lenmon:&          33.41\%\\ \hline
		
	\end{tabular}
	\label{tab:1}
\end{table*}

\begin{figure*}
	\centering
	\subfigure[Original samples]{
		\includegraphics[width=\linewidth]{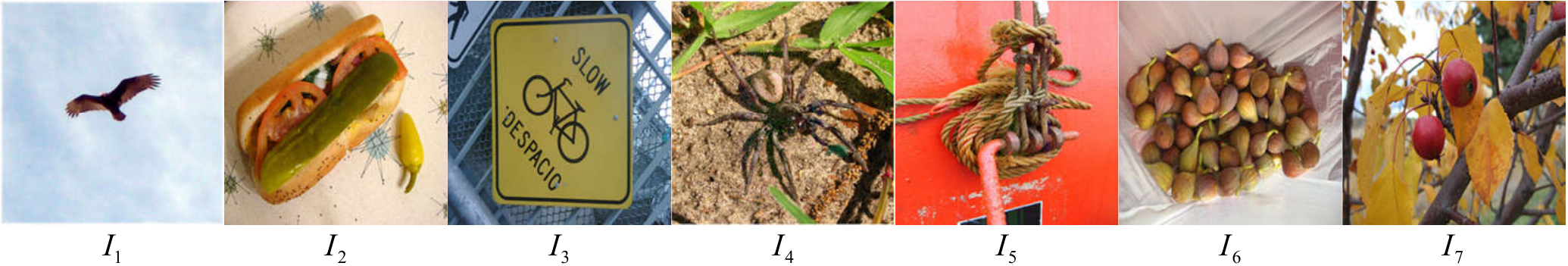}
	}
	\subfigure[Perturbation patterns and AEs generated by LMOA on ResNet-101]{
		\includegraphics[width=\linewidth]{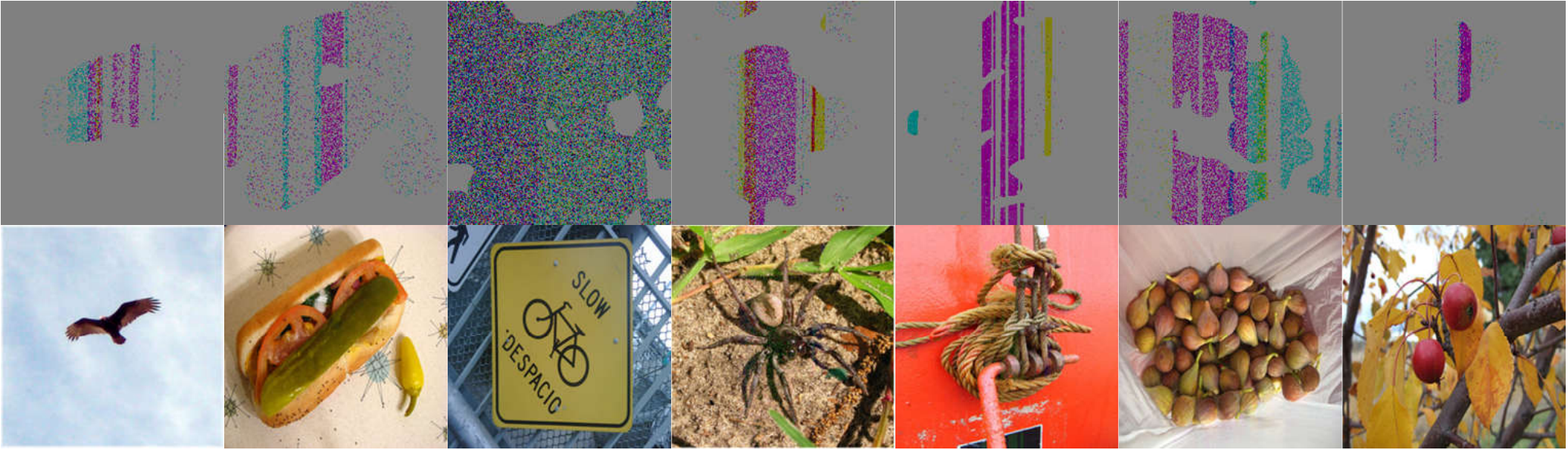}
	}
	\subfigure[Perturbation patterns and AEs generated by LMOA on Inception-v3]{
		\includegraphics[width=\linewidth]{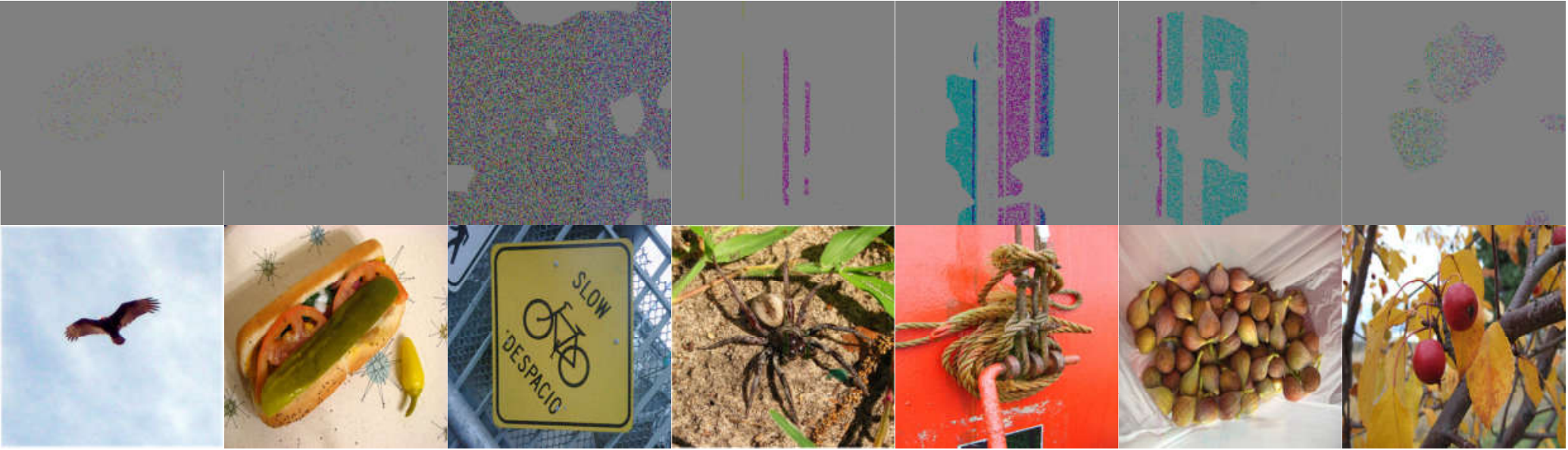}
	}
	\caption{Illustration of AEs obtained by the proposed LMOA.}
	\label{fig:4}
\end{figure*}

For MOEAs, the population size is fixed as $50$ for each image, and the number of maximum generation is set to $200$. The distribution indexes of SBX and PM are set to 20,  and the probabilities of crossover and mutation are set to $1.0$ and $1/d$, where $d$ is the dimension of decision variables for each adversarial attack task.

All the experiments are carried out on a PC with Intel Core i7-6700K 4.0GHz CPU, 48GB RAM, Windows 10, and Matlab R2018b with PlatEMO \citep{Tian2017}.

\subsection{Results and Analysis}
In this part, we first exhibit the performance of LMOA on the benchmark dataset. Then, we compare LMOA with state-of-the-art attacks. Finally, we also compare LMOA with a similar black-box attack approach that also draws on the idea of multi-objective optimization (we termed the method as MOEA-based attack).

\subsubsection{The performance of LMOA}
Fig. \ref{Pareto} shows the Pareto optimal solutions obtained by LMOA with seven images that randomly selected from the benchmark dataset. From the observations, we can conclude that the proposed LMOA can obtain the trade-off solutions between the two objectives including the confidence probability and $l_{2}$ norm. That is, the solution with a high confidence of the true label has a low $l_{2}$ norm, and vice versa. Note that, we select the perturbed image that successfully fools the target DNN while achieves the minimum $l_{1}$ norm as the final AE for each image.

\begin{table*}[tbp]
	\centering
	\renewcommand{\arraystretch}{1.5}
	\caption{Comparison on the classification accuracy, query count, and average $l_{2}$ norm of AEs between LMOA and state-of-the-art attacks.}
	\begin{tabular}{c|c|c|c|c}
		\hline
		Target model&                       Attack method&                Classification   accuracy (\%)&  Average $l_{2}$ norm &Query count  \\\hline
		\multirow{6}{*}{ResNet-101}&        N/A&                   77.10  &      N/A  &N/A\\ \cline{2-5}
		&      Boundary Attack \citep{brendel2017decision}&  30.00  &0.036  &125,000\\\cline{2-5}
		&      ZOO \citep{chen2017zoo}&  10.00        &4.82    & 264,170\\\cline{2-5}
		&      FGSM \citep{goodfellow2014explaining}&20.00          & 0.70  &N/A \\\cline{2-5}	
		&      FineFool \citep{chen2021finefool}&  2.89      &4.16    &N/A\\\cline{2-5}			
		&      ELAA \citep{guoelaa2021}&  1.02     &3.89    &N/A\\\cline{2-5}	
		&      \textbf{LMOA}&      \textbf{0.00}  &\textbf{3.96}&\textbf{10000}\\\hline

		\multirow{6}{*}{Inception-v3}&N/A      &78.40  &N/A&N/A \\\cline{2-5}
		&      Boundary Attack \citep{brendel2017decision}&  30.00  &0.19&125,000\\\cline{2-5}
		&      ZOO \citep{chen2017zoo}&   12.00       & 7.51   & 223,143\\\cline{2-5}
		&      FGSM \citep{goodfellow2014explaining}& 18.00         &0.41   &N/A \\\cline{2-5}	
		&      FineFool \citep{chen2021finefool}&0.05         &   3.21 &N/A\\\cline{2-5}
		&      ELAA \citep{guoelaa2021}&  0.04     &2.17   &N/A\\\cline{2-5}				
		&      \textbf{LMOA}&     \textbf{0.00}  &\textbf{4.06}&\textbf{10000}  \\\hline		
	\end{tabular}
	\vspace{1mm}
	\label{tab:black}
\end{table*}

Table \ref{tab:1} shows the classification results and confidence probabilities of the above seven original images and their corresponding AEs obtained with two DNNs. From the table, two remarks can be concluded as follows. Firstly, for the images that are correctly classified by the two DNNs, the proposed LMOA finally generates AEs that fool the models with high confidence probabilities (at least $22.9\%$, most of the results over $40\%$). Secondly, for the images that are misclassified by DNNs ($I_{1}$ and $I_{2}$ against Inception-v3), the proposed algorithm improves the confidence probabilities of the incorrect label (from $15.3\%$ to $70.98\%$ on $I_{1}$, $10.03\%$ to $59.93\%$ on $I_{2}$). Fig. \ref{fig:4} further exhibits these AEs with the proposed LMOA and two DNNs. We note that the value of perturbations includes both positive and negative values. For better presentation, the unmodified pixels in each image are exhibited in gray pixels. Besides, the brighter pixels indicate increases in their intensity, while darker pixels represent changes in the opposite direction of their intensity. As can be seen in Fig. \ref{fig:4}, the AEs generated by LMOA have similar visual perception compared with the original images. The reason is that the overall intensity of the perturbations is relatively low and the changes only emerge in part of the salient regions. Furthermore, even for the image that contains a simple background region with a sparse texture ($I_{1}$), the generated adversarial perturbation is still not easily perceptible. Therefore, the above evidence supports the analysis given in Section III, and it also verifies the effectiveness of the proposed method.

\subsubsection{Comparison with state-of-the-art attacks}
To validate the effectiveness of the proposed LMOA. We first compared five state-of-the-art attacks with LMOA. Table \ref{tab:black} shows the classification accuracy, query counts, and average l2 norm of AEs between LMOA and other attacks. From the table, we can see that 77.1\% of the benchmark images can be correctly classified by ResNet-101 and 21.6\% of the benchmark images are incorrectly classified by inception-v3. Moreover, we can see that LMOA achieves a great performance in all three metrics. Compared to the other five attacks, LMOA has a higher success rate and fewer queries, while generating fewer perturbations.

For classification accuracy, it can be seen from Table\ref{tab:black} that Boundary Attack has the highest classification accuracy among the five compared methods. It indicates that Boundary Attack has the lowest success rate on two target DNNs. Second, ZOO obtained a classification accuracy of 10\% and 12\% on two DNNs, respectively. It also requires about 220,000 queries, which seems to be more than the number of queries for the other black-box attacks. Finally, in terms of accuracy, LMOA achieves a performance similar to white-box attacks, even better than FineFool and ELAA. The main reason for the advantage of LMOA is that LMOA considers the spatial semantic information of images and combines it with the attention map to limit the range of perturbation generation. The experimental results also demonstrate that it is more effective to attack the region that the model is more concerned with.

\begin{figure*}
	\centering
	\subfigure[AE generated by the MOEA-based
	attack]{
		\includegraphics[width=\linewidth]{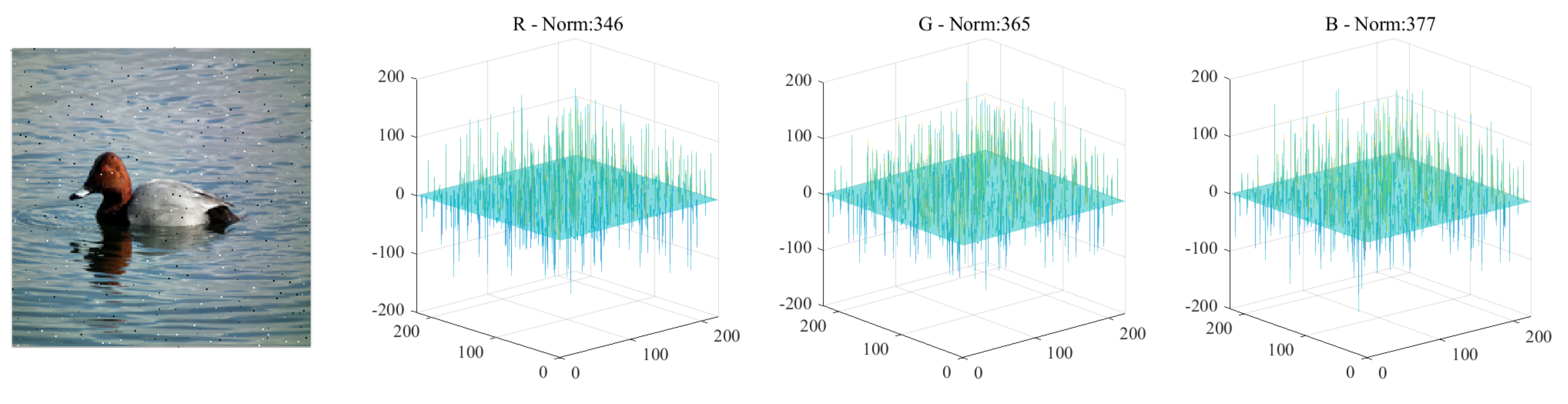}
	}
	\subfigure[AE generated by LMOA]{
		\includegraphics[width=\linewidth]{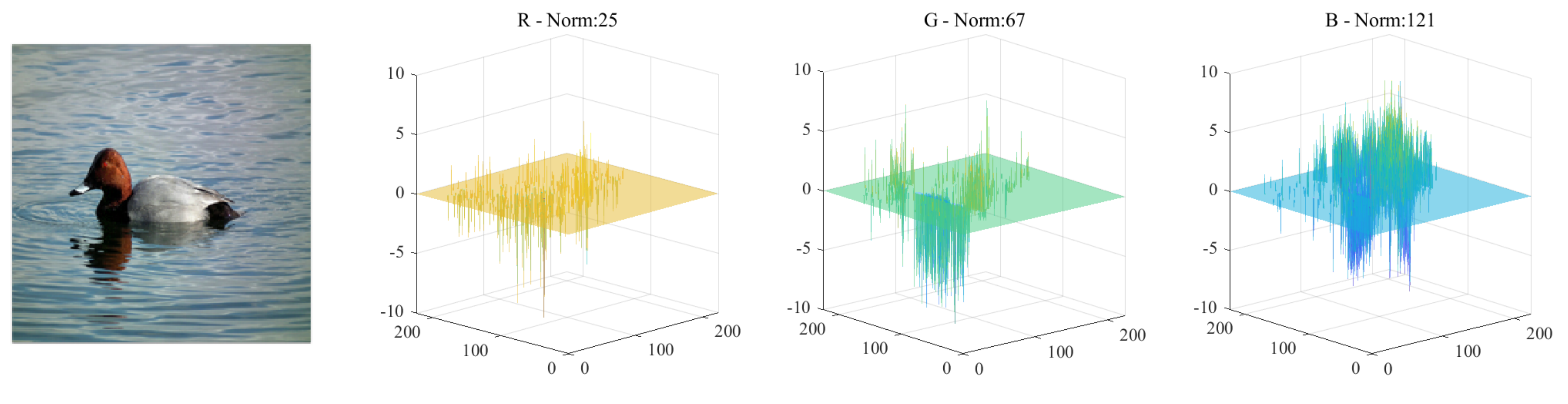}
	}
	\caption{The perturbations of the MOEA-based
		attack and the proposed LMOA.}
	\label{fig:5}
\end{figure*}

\begin{table}[htbp]
	\centering
	\renewcommand{\arraystretch}{1.5}
	\caption{Classification results and corresponding confidences of the original and AEs.}
	\begin{tabular}{c|c|c}
		\hline
		Target &                       Attack&                Classification\\
		model&  method &accuracy\\\hline
		\multirow{3}{*}{ResNet-101}&      N/A&                   100\% \\ \cline{2-3}
		&      with MOEA-based
		attack&  6.67\% \\\cline{2-3}
		&      with LMOA&      0\% \\\hline
		\multirow{3}{*}{Inception-v3}&    N/A&       86.67\% \\ \cline{2-3}
		&      with MOEA-based
		attack&  16.67\% \\\cline{2-3}
		&      with LMOA&      0\% \\\hline
		
	\end{tabular}
	\label{tab:2}
\end{table}

For the average $ l2 $ norm, although Boundary Attack has the smallest $ l_2 $ norm, it performs worse than other attacks in terms of both query count and classification accuracy. Similarly, ZOO requires nearly 260,000 queries to achieve 10\% classification accuracy. lMOA requires only 10,000 queries, and it also achieves 0\% classification accuracy while maintaining an acceptable $ l_2 $ norm. Note that LMOA has a slightly inferior average $l_2$ norm than Finefool and ELAA for the inception-v3 model. This is explained by the fact that white-box attacks can guide perturbation generation based on gradient information, resulting in smaller perturbations. It is still a challenge to generate smaller perturbations under black-box attacks.

\subsubsection{Comparison with MOEA-based attack}

\begin{figure}[htbp]
	\centering
	\subfigure[AEs generated by the MOEA-based
	attack]{
		\includegraphics[width=3in]{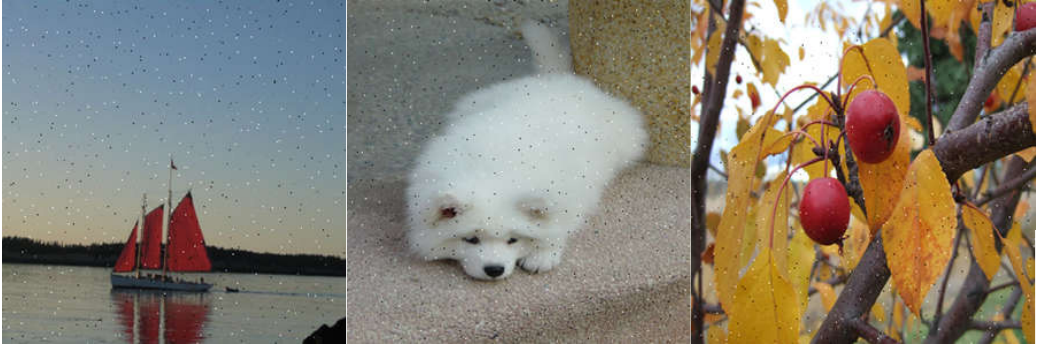}
	}
	\subfigure[AEs generated by the proposed LMOA]{
		\includegraphics[width=3in]{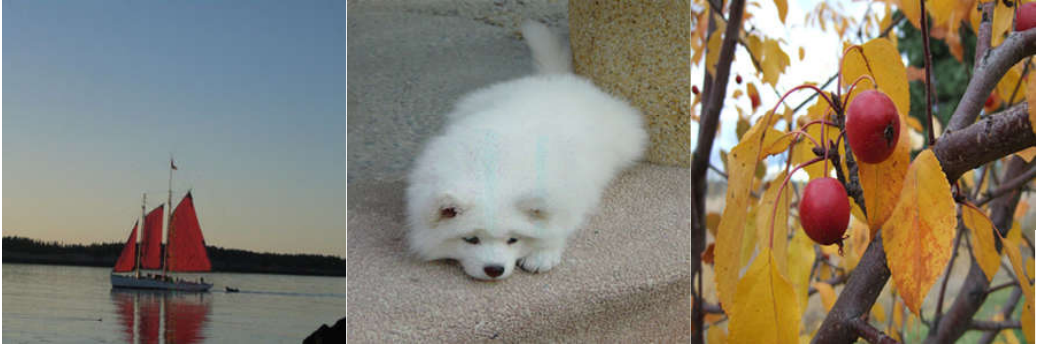}
	}
	
	\caption{The visual comparison of the AEs generated by the MOEA-based attack and the proposed LMOA.}
	\label{fig:6}
\end{figure}

Table \ref{tab:2} compares the proposed LMOA and another MOEA-based black-box attack method \citep{suzuki2019adversarial}, which adopts block-division method and formulates a MOP solved by MOEA/D \citep{Qingfu2007MOEA}. From the table, we can first observe that all of the images are correctly classified by ResNet-101, whereas $4$ of $30$ images are misclassified by Inception-v3. After performing the MOEA-based attack, only $2$ and $5$ images are correctly classified by two DNNs, respectively. By contrast, all the perturbed images generated by the proposed LMOA successfully attack the two models. Fig. \ref{fig:6} also visualized attack results obtained by the two methods. From the figure, we show that the MOEA-based attack adds some visible noises to each entire image, which can be easily captured by the human vision. For LMOA, the perturbations are much more difficult to be perceived compared with the MOEA-based attack. The advantage is attributed to the usage of the attention mechanism and large-scale MOEA.

Fig. \ref{fig:5} further compares the adversarial attacks performed by the two methods. Similarly, the top row shows the generated AE and $l_2$ norm value on each channel obtained with the MOEA-based attack. The bottom row depicts the corresponding results of the proposed LMOA. From the figure, two observations are summarized as follows. Firstly, in terms of the region of perturbations, the MOEA-based attack attacks the entire image, resulting in easily perceptible perturbations. In contrast, this paper only perturbs the pixels in the salient region, which results in more sparse and smooth attacks. Secondly, the perturbation intensity of the bottom image is significantly smaller than that of the top one. This is also supported by the $l_2$ norm values of three channels. The fact indicates that the adversarial attack of LMOA is more effective than the MOEA-based attack on high-resolution images.

\begin{figure}[tbp]
	\centering
	\includegraphics[width=\linewidth]{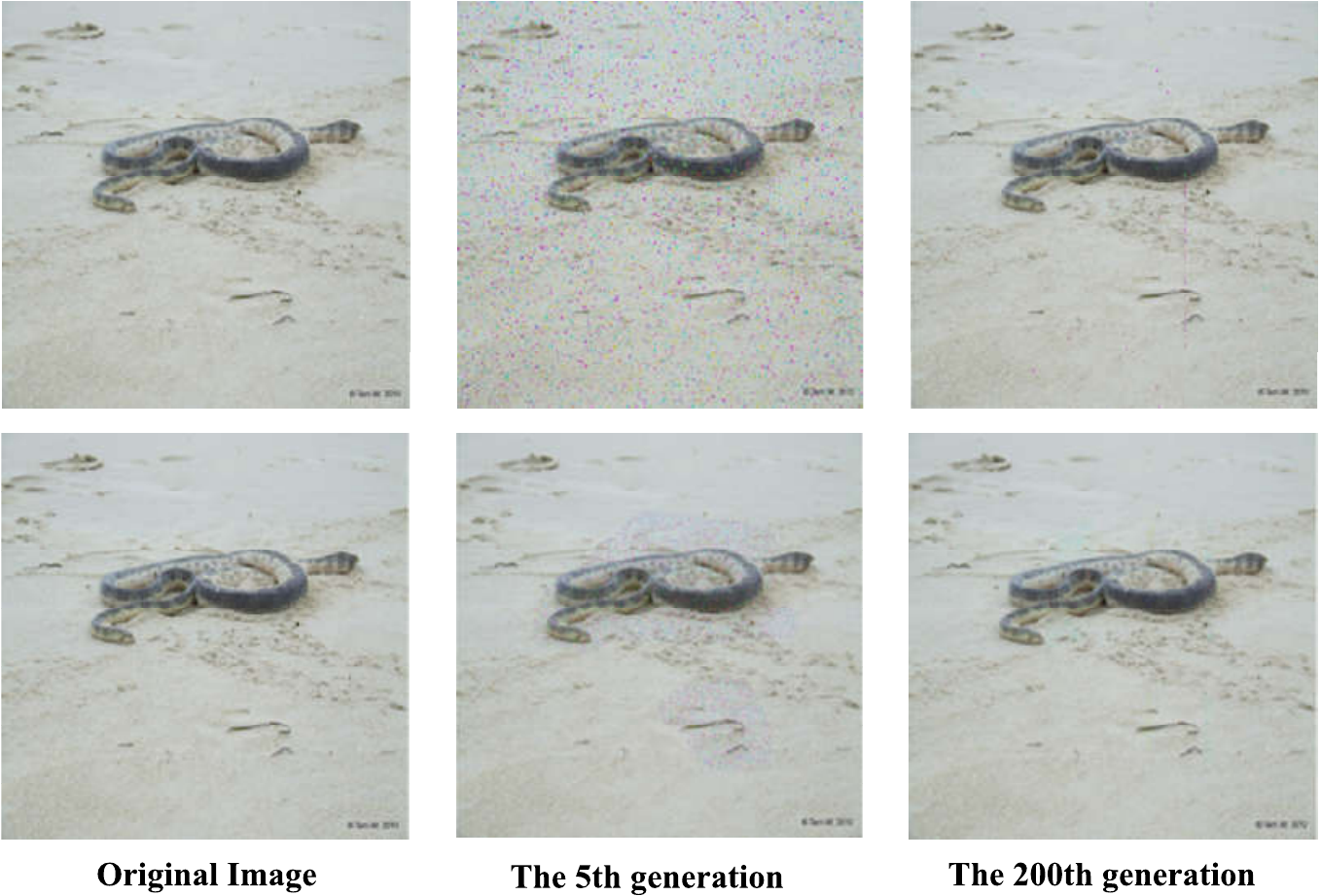}
	\caption{Illustration of the evolving processes of MOEA/PSL without or with the attention mechanism.}
	\label{fig:7}
\end{figure}

\begin{figure}[htbp]
	\centering
	\subfigure[The Pareto optimal solutions obtained by NSGA-II]{
		\includegraphics[width=\linewidth]{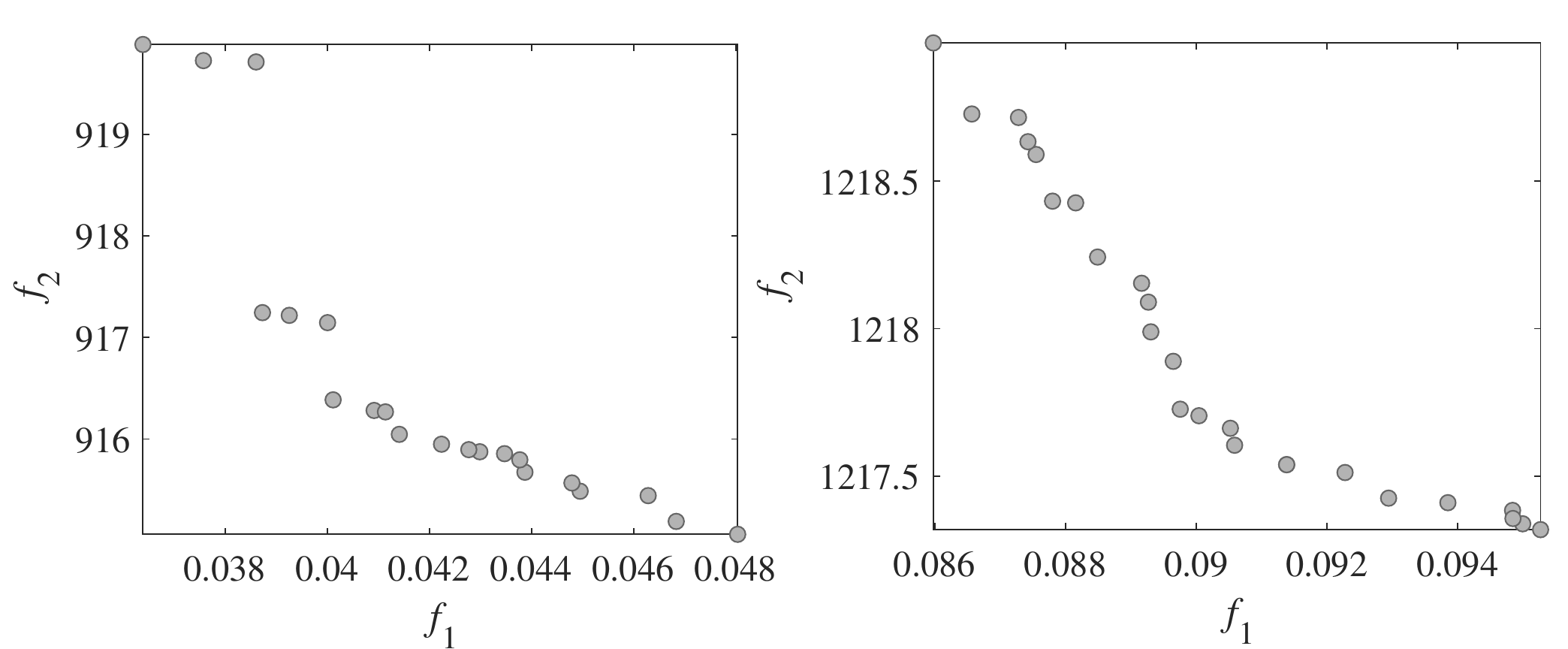}
	}
	\subfigure[The Pareto optimal solutions obtained by SPEA2]{
		\includegraphics[width=\linewidth]{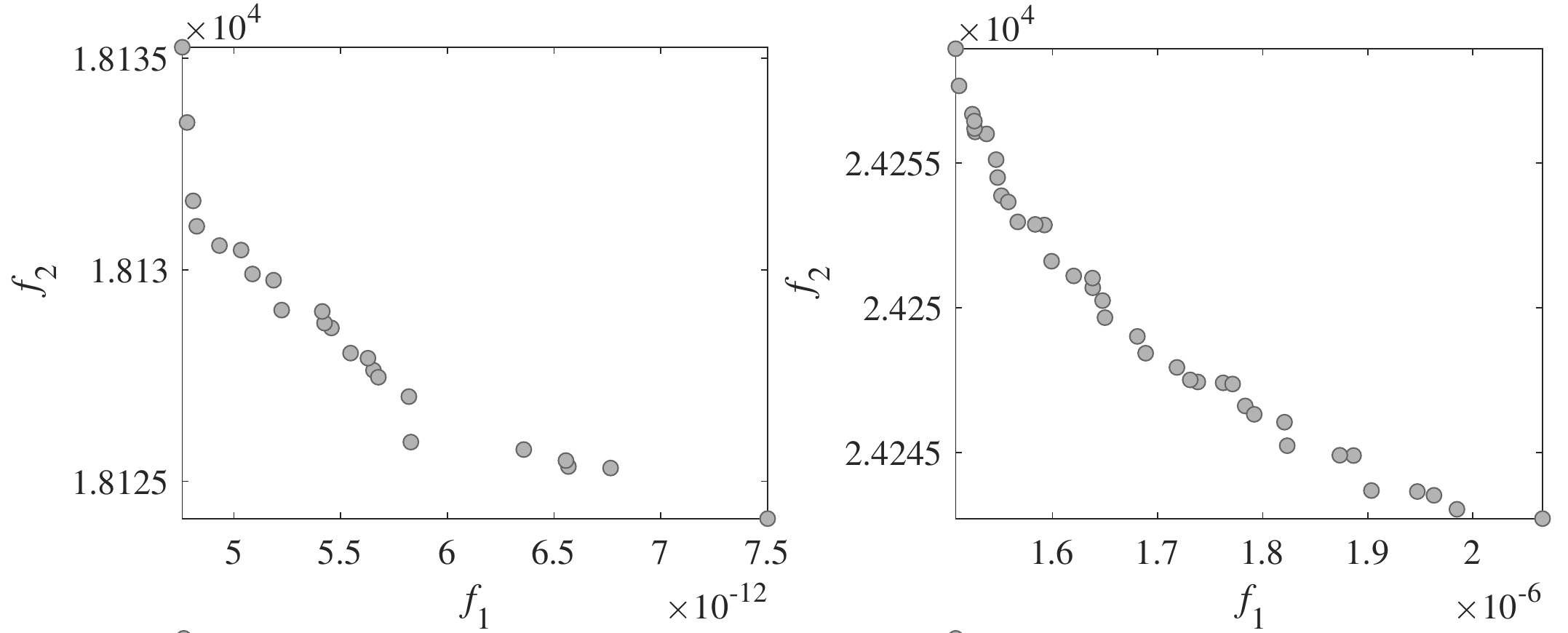}
	}
	\subfigure[The Pareto optimal solutions obtained by MOEA/PSL]{
		\includegraphics[width=\linewidth]{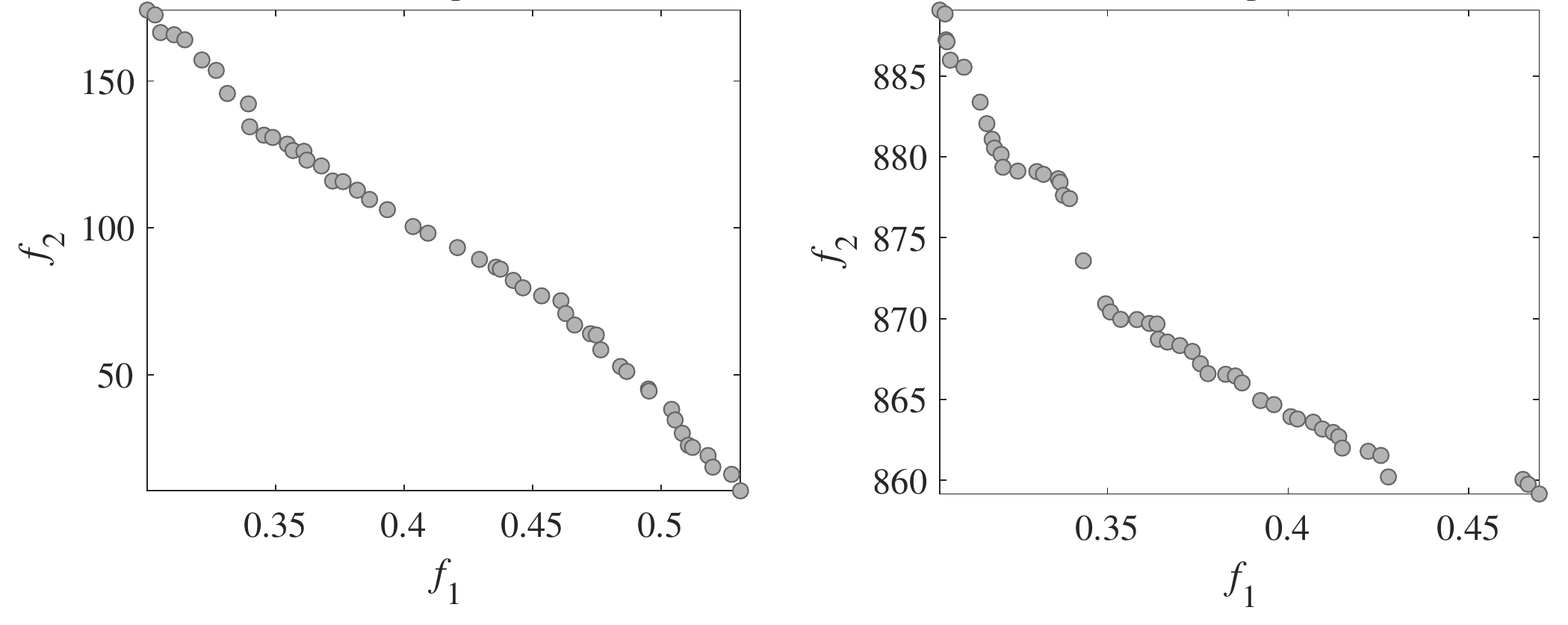}
	}
	\caption{The Pareto optimal solutions obtained by three MOEAs on ResNet-101 and Inception-v3 with two randomly selected images.}
	\label{fig:3pareto}
\end{figure}

\subsection{Ablation Study}
In this part, we investigate the effectiveness of three components in LMOA through an ablation study. 

\subsubsection{Effectiveness of using attention guided search}
The existing study has shown that attacking the salient region is more effective than attacking the background region \citep{dong2020robust}. To verify the point, they performed the experiments, where the perturbations are imposed on the background and foreground regions separately. It was found that images that changed the foreground region can still fool the classifier, while another one failed. The above experiment result implies that adding perturbations to the salient region can get better attack performance. Moreover, we investigate the effectiveness of the attention mechanism when searching for the optimal adversarial attacks with MOEAs in this part. In other words, we attempt to explain the advantage of using the attention mechanism for black-box adversarial attacks with a new viewpoint.

We randomly pick an image from the benchmark dataset and perform the search by MOEA/PSL with the following two manners. The one is attacking the image without the guide of the attention mechanism, \emph{i.e.,} the search space of the perturbation is the entire image. The other one aims at searching for the optimal perturbation guided by the attention map, which means that the region of perturbation is limited by the attention map.

Fig. \ref{fig:7} compares the results of two search paradigms. The original image, the AE obtained with 5 generations, and the AE obtained with 200 generations are shown from left to right in order. It can be seen that the AE obtained with the attention mechanism guided search contains significantly fewer noise than another one, regardless of the number of generations. The fact shows that restricting the perturbation region can facilitate the convergence of MOEA/PSL, which provides a more effective and invisible adversarial attack on the selected image. We also note that the reduction of the search space can greatly alleviate the computational cost of adversarial attacks with large-scale MOEAs.

\subsubsection{Effectiveness of using Large-scale MOEA}
This paper employs one of the MOEAs that are specially designed for tackling large-scale MOPs, due to the high dimension of the optimization problem formulated for the high-resolution images. To verify the necessity of this technique, we further investigate the performance of the other two popular MOEAs, including NSGA-II \citep{Deb2002A}, and SPEA2 \citep{Zitzler2001}, on the proposed MOP.

\begin{figure}[tbp]
	\centering
	\subfigure[AEs generated by LMOA with ResNet-101]{
		\includegraphics[width=\linewidth]{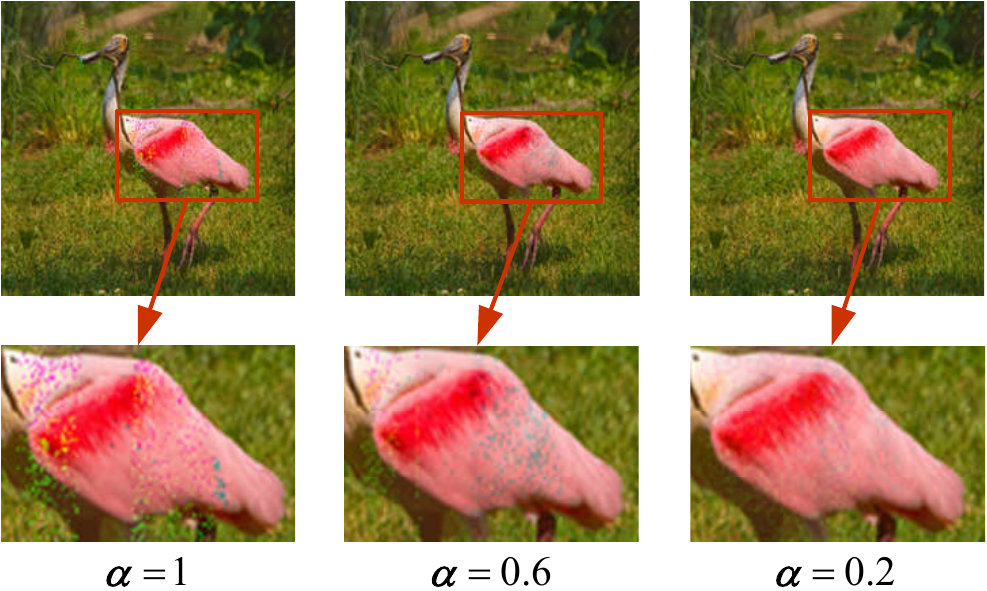}
	}
	\subfigure[AEs generated by LMOA with Inception-v3]{
		\includegraphics[width=\linewidth]{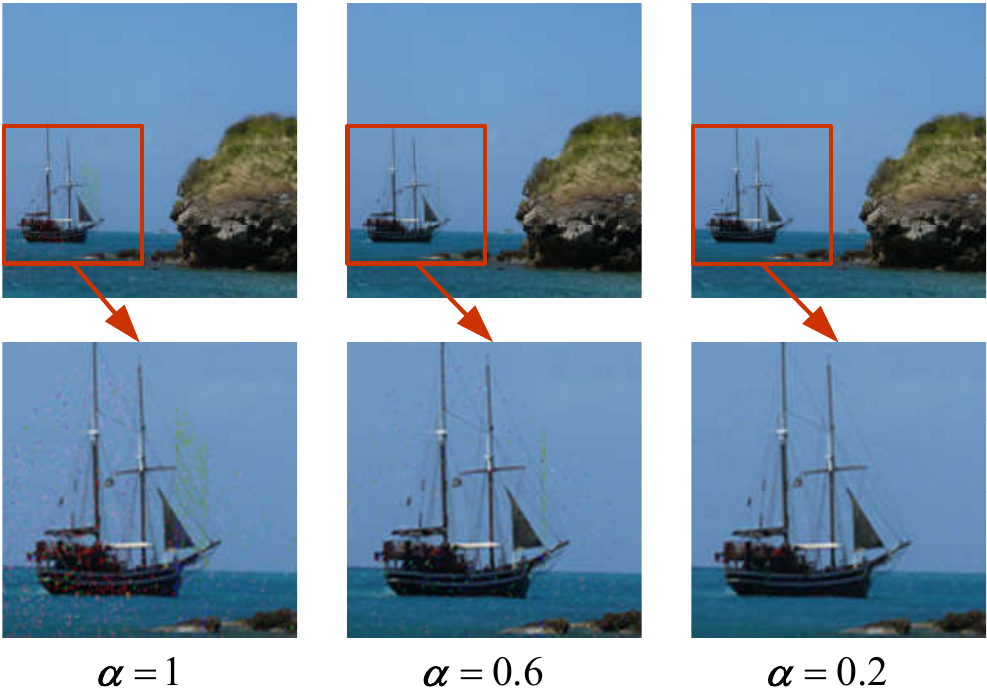}
	}
	\caption{Illustration of the sensitivity of the parameter $\alpha$ involved in the improved population initialization.}
	\label{fig:8}
\end{figure}

Fig. \ref{fig:3pareto} compares the Pareto optimal solutions obtained with three MOEAs on two DNNs, respectively. We note that using each of the three MOEAs can successfully attack the images in the experiment.  Although both NSGA-II and SPEA2 are able to generate AEs with low confidence probabilities with respect to the true label, they fail to find out the AEs with acceptable $l_{2}$ norm values, especially SPEA2. Compared with the above two MOEAs, MOEA/PSL can search for the perturbations with lower intensities on each image, while accomplishing the adversarial attack. This result may be explained by the fact that traditional MOEAs may struggle to the large-scale MOPs, and they are not very suitable for attacking high-resolution images under the black-box settings. The above observations also confirm the effectiveness of using MOEAs tailored for large-scale MOPs in this paper.

\subsubsection{Effectiveness of the improved population initialization}
As described in Section III, the initialized population is required to be multiplied with a small number $\alpha$ as the candidate AEs should maintain a high similarity with the original image. As known, a suitable initialization may help to result in better convergence of MOEAs. In this part, we will show that the improved population initialization can greatly influence the search results of LMOA.

Fig. \ref{fig:8} shows a comparison among the generated AEs with different values of $\alpha$. It is clear that the AEs without the improved population initialization is much easier to be perceived. By contrast, the AEs with $\alpha=0.6$ and $\alpha=0.2$ achieves better visual masking. The observations indicate that selecting a proper $\alpha$ can generate an imperceptible AE while ensuring the success rate. Note that, the value of $\alpha$ can affect the success rate of the attack. An extremely small $\alpha$ is unfavorable to generate AEs that misclassify the target network. In this paper, we use $\alpha = 0.2$ for every image.

\section{Conclusions}
As pointed by previous literature, traditional EA-based black-box attacks may struggle to tackle high-resolution images. On the other hand, existing black-box attacks usually focus on attacking the entire image, which results in visually more perceptible AEs. In this paper, we propose a novel attention-guided black-box adversarial attack, where the adversarial perturbations are only added to several pixels in the salient region. Besides, MOEA/PSL is used to search for the optimal perturbation, and the algorithm is used to solve MOPs, where the Pareto optimal solutions are sparse. Experimental results show that the proposed LMOA can perform effective black-box attacks against high-resolution images (with nearly 100\% success rate and high visual quality). Moreover, comparing with the MOEA-based attack,  the proposed LMOA is more suitable to addressing high-resolution images from ImageNet dataset, due to the usage of large-scale MOEA and attention-guided search.

Future works of this study will include the following respects. First, improving AEs generation by considering more characteristics of the image and dimension reduction techniques is a feasible way. Secondly, alleviating the computation cost of LMOA is another interesting topic. Thirdly, it is necessary to consider the cross-model generalization of EA-based black-box attacks.

\section*{Acknowledgment}
This research work is partly supported by National Natural Science Foundation of China No.62172001.

\bibliographystyle{IEEEtran}
\bibliography{manuscript.bib}

\end{document}